\documentclass[letterpaper, 10pt, conference]{ieeeconf}

\IEEEoverridecommandlockouts                              

\usepackage{graphicx, amsmath}
\usepackage{multirow}
\usepackage{cite, url, color}
\usepackage{balance}
\usepackage{epstopdf}

\title{\LARGE \bf Social Attention: Modeling Attention in Human Crowds}

\author{Anirudh Vemula, Katharina Muelling and Jean
  Oh
  \thanks{A. Vemula, K. Muelling and J. Oh are with the Robotics
    Institute, Carnegie Mellon University, Pittsburgh, Pennsylvania,
    USA 15213.  Email: \texttt{\{avemula1, katharam, jeanoh\}@cs.cmu.edu}}%
}

\begin{document}

\maketitle
\newcommand{\xhat}{\hat{x}}
\newcommand{\yhat}{\hat{y}}
\newcommand{\graph}{\mathcal{G}}
\newcommand{\vertices}{\mathcal{V}}
\newcommand{\edges}{\mathcal{E}}
\newcommand{\feature}{\mathbf{x}}
\newcommand{\unitx}{x}
\newcommand{\unity}{y}
\newcommand{\rnn}{\mathbf{R}}
\newcommand{\RNN}{\text{RNN}}
\newcommand{\spatial}{\text{spatial}}
\newcommand{\temporal}{\text{temporal}}
\newcommand{\node}{\text{node}}
\newcommand{\note}[1]{\textcolor{red}{#1}}
\newcommand{\concat}{\text{concat}}
\newcommand{\prob}{\text{P}}
\newcommand{\obs}{\text{obs}}
\newcommand{\pred}{\text{pred}}


\begin{abstract}
Robots that navigate through human crowds need to be able to plan safe, efficient, and human predictable trajectories. This is a particularly challenging problem as it requires the robot to predict future human trajectories within a crowd where everyone implicitly cooperates with each other to avoid collisions. Previous approaches to human trajectory prediction have modeled the interactions between humans as a function of proximity. However, that is not necessarily true as some people in our immediate vicinity moving in the same direction might not be as important as other people that are further away, but that might collide with us in the future. In this work, we propose \textit{Social Attention}, a novel trajectory prediction model that captures the relative importance of each person when navigating in the crowd, irrespective of their proximity. We demonstrate the performance of our method against a state-of-the-art approach on two publicly available crowd datasets and analyze the trained attention model to gain a better understanding of which surrounding agents humans attend to, when navigating in a crowd.
\end{abstract}


\section{Introduction}
\label{sec:introduction}

Robots are envisioned to coexist with humans in unscripted environments and accomplish a diverse set of objectives. Towards this goal, navigation is an essential task for the autonomous mobile robot. This requires the mobile robot to navigate human crowds in not just a safe and efficient manner, but also in a socially compliant way, i.e., the robot has to collaboratively avoid collisions with surrounding humans and alter its path in a human-predictable manner.  To achieve this, the robot needs to accurately predict the future trajectories of humans within the crowd and accordingly plan its own path.

Early works in the domain of social robot navigation have modeled individual human motion patterns in crowds to predict future trajectories as in~\cite{thompson09, bennewitz05, large04}. However, as shown in~\cite{trautman10}, such independent modeling does not capture the complex and subtle interactions between humans in the crowd and the resulting path for the robot is highly suboptimal. For the robot to navigate in a socially compliant way, it is key to capture human-human interactions observed in a crowd.

More recent approaches such as~\cite{trautman10, VemulaMO17, Alahi-2016-ID2} model the joint distribution of future trajectories of all interacting agents through a spatially local interaction model. Such a joint distribution model is capable of capturing the dependencies between trajectories of interacting humans, and results in socially compliant predictions. However, these approaches assume that only humans in a local neighborhood affect each other's motion, which is not necessarily true in real crowd scenarios. For example, consider a long hallway with two humans moving at both ends towards each other. If both of them were walking, such an assumption holds as they don't influence each other over such long distance. However, if one of them starts running, the other person adapts his own motion to avoid collision before the runner enters his local neighborhood. This observation leads us to the insight that human-human interactions in crowd are not just dependent on relative distance, but also on other features such as velocity, time-to-collision~\cite{karamouzas2014universal}, acceleration and heading.

\begin{figure}[t]
  \centering
  \includegraphics[width=0.9\linewidth]{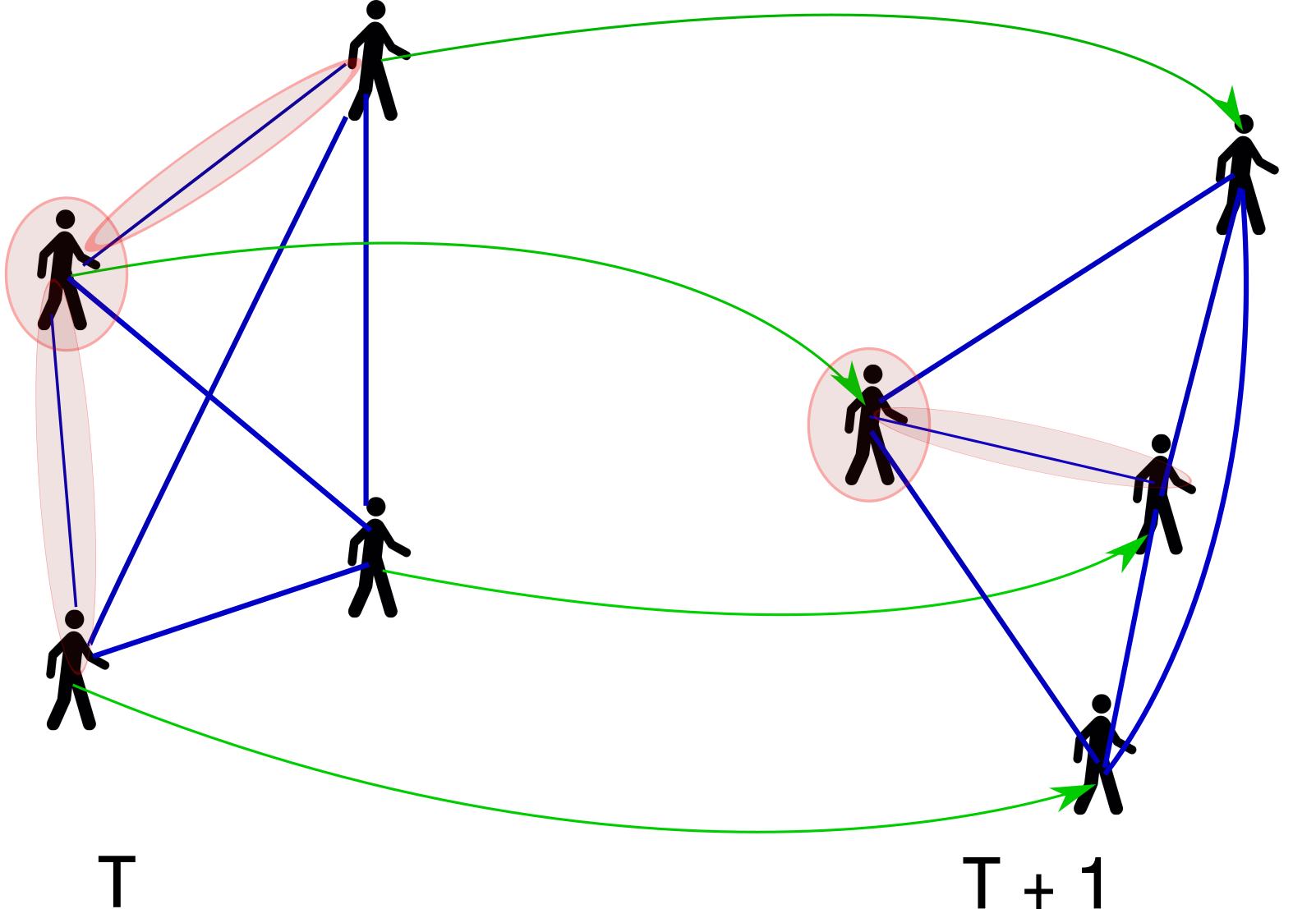}
  \caption{Humans, when navigating a crowd, pay attention to only a subset of surrounding agents at each time-step. In this work, we seek to learn such an attention model over surrounding agents to predict trajectories of all agents in the crowd more accurately by capturing subtle human-human interactions.}
  \label{fig:intro}
  \vspace*{-0.5cm}
\end{figure}

In this work, we propose an approach that addresses this observation through a novel data-driven architecture for predicting future trajectories of humans in crowds. As a foremost step towards achieving socially acceptable robot navigation, we focus on the problem of human trajectory prediction in a crowd. 
We use a feedforward, fully differentiable, and jointly trained
recurrent neural network (RNN) mixture to model trajectories of all
humans in the crowd, addressing both spatial and temporal aspects of
the problem. The human-human interactions are modeled using a soft
attention model over \textit{all} humans in the crowd, thereby not
restricting the approach with the local neighborhood assumption
(Figure \ref{fig:intro}). The resulting model captures the influence of each person on the other, the nature of their interaction and predicts their future trajectories. Finally, we demonstrate that our model, \textit{Social Attention}, is capable of predicting human trajectories more accurately than the state-of-the-art approach on two publicly available real world crowd datasets. We also analyze the trained attention model to understand the nature of human-human interactions learned from the crowd datasets.



\section{Problem Definition}
\label{sec:problem-definition}
In this paper, we deal with the problem of human trajectory prediction in crowded spaces. We assume that each scene is preprocessed to track pedestrians in the crowd and obtain their spatial coordinates at successive time-steps. Note that, across time-steps pedestrians enter and leave the scene, with varying length trajectories. Let $(x^t_i, y^t_i)$ represent the spatial location of agent $i$ at time-step $t$.

Following a similar notation as \cite{Alahi-2016-ID2}, our problem can
be formulated as: Given spatial locations $\{(x_i^t, y_i^t)\}$ for
agents $i = 1, 2, \cdots, N$  from time-steps $t = 1, \cdots,
T_{\obs}$, predict their future locations $\{(\xhat_i^t, \yhat_i^t)\}$
from $t = T_{\obs}+1, \cdots, T_{\pred}$.

\section{Related Work}
\label{sec:related-work}

Our work is relevant to past literature in the domain of modeling human interactions for navigation, human trajectory prediction and spatio-temporal models.

\subsection{Modeling Human Interactions for Navigation}
\label{sec:model-human-inter}
To predict future behavior of pedestrians in crowds, we need to model interactions between pedestrians accurately. An early work by~\cite{helbing95} proposed \textit{Social Force}, which models motion of pedestrians using attractive forces that guide them towards the destination, and repulsive forces that ensure collision-avoidance. Subsequently, several approaches~\cite{johansson07, Mehran-2009-ID10} have extended the social forces model by fitting the parameters of the force functions to observed crowd behavior. Using attractive and repulsive forces based on relative distances, the social forces model can capture simple interactions but can't model complex crowd behavior such as cooperation, as shown in~\cite{Alahi-2016-ID2}.

A pioneering work by~\cite{hall63} introduced a theory on human proximity relationships which has been used in potential field based methods such as~\cite{pradhan11} to model human-human interactions in crowds for robot navigation. The proximity-based model effectively captures reactive collision-avoidance but does not model human-human and human-robot cooperation. However, models of cooperation are essential for safe and efficient robot navigation in dense crowds. {As shown by~\cite{trautman10},} lack of cooperation leads to the \textit{freezing robot problem} where the robot believes there is no feasible path in the environment, despite the existence of several feasible paths.

{More recently, the use of \textit{Interacting Gaussian Processes} (IGP) was proposed by~\cite{trautman10}}
to model the joint distribution of trajectories of all interacting agents in the crowd using Gaussian Processes with a handcrafted interaction potential term. The potential term captures interactions based on the relative distances of humans in the crowd and results in a probabilistic model that has been shown to capture joint collision avoidance behavior. This has been extended in~\cite{VemulaMO17} by replacing the handcrafted potential term with a {locally} trained interaction model based on occupancy grids. However, these approaches model interactions based on relative distances and orientations, ignoring other features such as velocity and acceleration.

Finally, the works of~\cite{kuderer12, kretzschmar16} explicitly model human-human and human-robot interactions and jointly predict the trajectories of all agents, using feature-based representations. They use \textit{maximum entropy inverse reinforcement learning} (IRL) to learn a distribution of trajectories that results in crowd-like behavior. Features used such as clearance, velocity, and group membership are carefully designed. However, their approach has only been tested in scripted environments with no more than four humans and due to the feature-based joint modeling, it scales poorly with the number of agents considered. Very recently,~\cite{pfeiffer16} extended this approach to unseen and unstructured environments using a receding horizon motion planning approach.

\subsection{Human Trajectory Prediction}
\label{sec:human-traj-pred}

In the domain of video surveillance, human trajectory prediction is a significant challenge. {The approaches by}~\cite{kim11, joseph2011bayesian} learn motion patterns of pedestrians in videos using Gaussian Processes and cluster observed trajectories into patterns. These motion patterns capture navigation behavior such as static obstacle avoidance, but they ignore human-human interactions.
IRL has also been used for activity forecasting in~\cite{kitani2012activity} to predict future trajectories of pedestrians by inferring traversable regions in a scene by modeling human-space interactions using semantic scene information. However, interactions between humans are not modeled. More recently,~\cite{Alahi-2016-ID2} used Long Short-Term Memory networks (LSTM) to model the joint distribution of future trajectories of interacting agents. This work has been extended in~\cite{varshneya2017human, bartoli2017context} to include static obstacles in the model in addition to dynamic agents. However, these approaches assume that only the dynamic agents in a local discretized neighborhood of a pedestrian affect the pedestrian's motion. As shown in Section \ref{sec:introduction}, this is not necessarily true and in our work, we do not make such an assumption. The authors would also like to point out a very recent work~\cite{fernando2017soft+} who also consider all agents in the environment, rather than just the local neighborhood, using attention. However, the attention used is hard-wired based on proximity rather than being learned from data.

\subsection{Spatio-Temporal Models}
\label{sec:spat-temp-models}

In this paper, we formulate the task of human trajectory prediction
using spatio-temporal graphs. Spatio-temporal graphs have nodes that
represent the problem components and edges that capture
spatio-temporal interactions between the nodes.  This spatio-temporal
formulation finds applications in robotics and computer
vision,~\cite{fragkiadaki2015recurrent, sun2013active,
  jain2015car}. Traditionally, graphical models such as Conditional
Random Fields are used to model such problems,~\cite{li2008key,
  koppula2016anticipating,
  zhang2014overtaking}. Recently,~\cite{jain2016structural} introduced
\textit{Structural RNN}(S-RNN), a rich RNN mixture that can be jointly
trained to model dynamics in spatio-temporal tasks. This has been
successfully applied to diverse tasks such as modeling human motion
and driver maneuver anticipation. In this paper, we will use a variant
of S-RNN.



\section{Approach}
\label{sec:approach}

Humans navigate crowds by adapting their own trajectories based on the motion of others around them. It is assumed in~\cite{Alahi-2016-ID2, VemulaMO17, varshneya2017human, bartoli2017context} that this influence is spatially local, i.e., only spatial neighbors influence the motion of a human in the crowd. But as shown in Section~\ref{sec:introduction}, this is not necessarily true and other features such as velocity, acceleration and heading play an important role, enabling agents who are not spatially local to influence a pedestrian's motion. In this work, we aim to model the influence of all agents in the crowd by learning an attention model over the agents. In other words, we seek to answer the question: \textit{Which surrounding agents do humans attend to, while navigating a crowd?} Our hypothesis is that the representation of trajectories learned by our model enables us to effectively reason about the importance of surrounding agents better than only considering spatially local agents.

As argued in Section~\ref{sec:introduction}, to model interactions among humans, we cannot predict future locations of each human independently. Instead, we need to jointly reason across multiple people and couple their predictions so that interactions among them are captured. Towards this goal, we use a feedforward, fully differentiable, and jointly trained RNN mixture that predicts both their future locations and captures human-human interactions. 
Our approach builds on the architecture proposed in~\cite{jain2016structural} for this purpose.

\subsection{Spatio-Temporal Graph Representation}
\label{sec:spat-temp-graph}

We use a similar spatio-temporal graph (st-graph) representation as~\cite{jain2016structural} with $\graph = (\vertices, \edges_S, \edges_T)$, where $\graph$ is the st-graph, $\vertices$ is the set of nodes, $\edges_S$ is the set of spatial edges and $\edges_T$ is the set of temporal edges. Note that the graph $(\vertices, \edges_S)$ is unrolled using $\edges_T$ to form $\graph$. Hence, in the unrolled st-graph, different nodes at the same time-step are connected using edges $\edges_S$ whereas same nodes at adjacent time-steps are connected using edges $\edges_T$. For more details on general st-graph representation, we refer the reader to~\cite{jain2016structural}.

\begin{figure*}[!ht]
  \centering
  \includegraphics[width=0.8\linewidth]{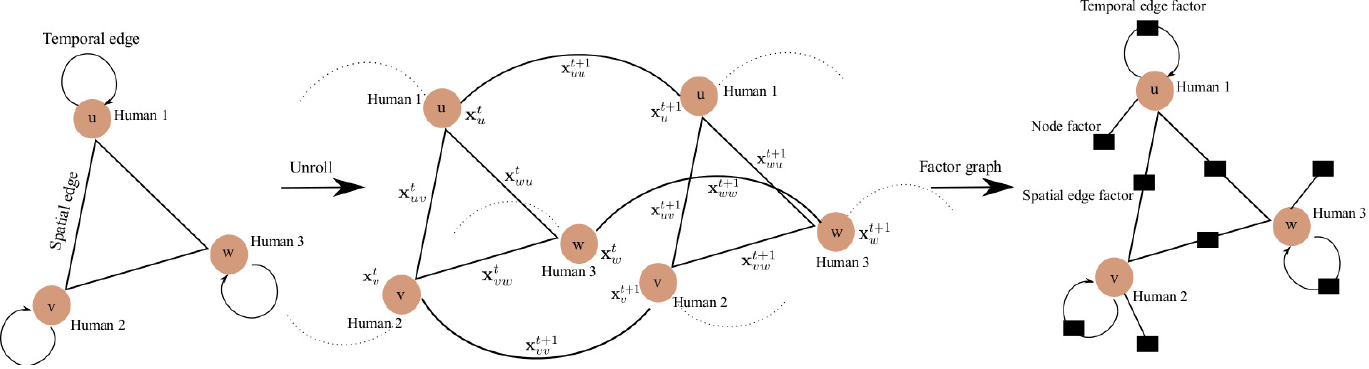}
  \caption{Example st-graph, unrolled st-graph for two time steps and corresponding factor graph}
  \label{fig:stgraph}
  \vspace*{-0.5cm}
\end{figure*}

In this work, we formulate the problem of human trajectory prediction as a spatio-temporal graph. The nodes of the st-graph represent the humans in the crowd, the spatial edges connect two different humans at the same time-step, and temporal edges connect the same human at adjacent time-steps. The spatial edges aim to capture the dynamics of relative orientation and distance between two humans, and temporal edges capture the dynamics of the human's own trajectory. The feature vector associated with node $v$ at time-step $t$ is $\feature_v^t = (x_v^t, y_v^t)$, the spatial location of the corresponding human. The feature vector associated with a spatial edge $(u, v) \in \edges_S$ at time-step $t$ is $\feature_{uv}^t = (\unitx_{uv}^t, \unity_{uv}^t)$, the vector from location of $u$ at time $t$ to location of $v$ at $t$ (encoding the relative orientation and distance). Similarly, the feature vector associated with a temporal edge $(u, u) \in \edges_T$ at time-step $t$ is $\feature_{uu}^t = (\unitx_{uu}^t, \unity_{uu}^t)$, the vector from location of node $u$ at $t-1$ to its location at $t$. The corresponding st-graph representation (with the unrolled st-graph) is shown in Figure \ref{fig:stgraph}.

The factor graph representation of the st-graph associates a factor function for each node and a pairwise factor function for each edge in the graph, as shown in Figure \ref{fig:stgraph}. At each time-step, the factors in the st-graph observe node/edge features and perform some computation on those features. Each of these factors have parameters that need to be learned. In our formulation, all the nodes share the same factor, giving the model scalability to handle more nodes (in dense crowds) without increasing the number of parameters. For similar reasons, all spatial edges share a common factor and all temporal edges share the same factor function. Note that the factor for spatial edges and temporal edges are different, as they capture different aspects of the trajectories. This kind of parameter sharing is necessary to generalize across scenes with varying number of humans, and keeps the parameterization compact.

\subsection{Model Architecture}
\label{sec:model-architecture}

The factor graph representation lends itself naturally to the S-RNN architecture~\cite{jain2016structural}. {We represent each factor with an RNN}. Hence, for each of the node factors we have nodeRNNs $\{\rnn_v\}$ and for each of the edge factors we have edgeRNNs $\{\rnn_{uv}\}$. Note that all the nodeRNNs, spatial edgeRNNs and temporal edgeRNNs share parameters among themselves. The spatial edgeRNNs model the dynamics of human-human interactions in the crowd and the temporal edgeRNNs model the dynamics of individual motion of each human in the crowd. The nodeRNNs use the node features and hidden states from the neighboring edgeRNNs to predict the future location of the node at the next time-step. { We would like to emphasize that since we share the model parameters across all nodes and edges, the number of parameters is independent of the number of pedestrians at any given time}.

\begin{figure*}[!ht]
  \centering
  \includegraphics[width=0.8\linewidth]{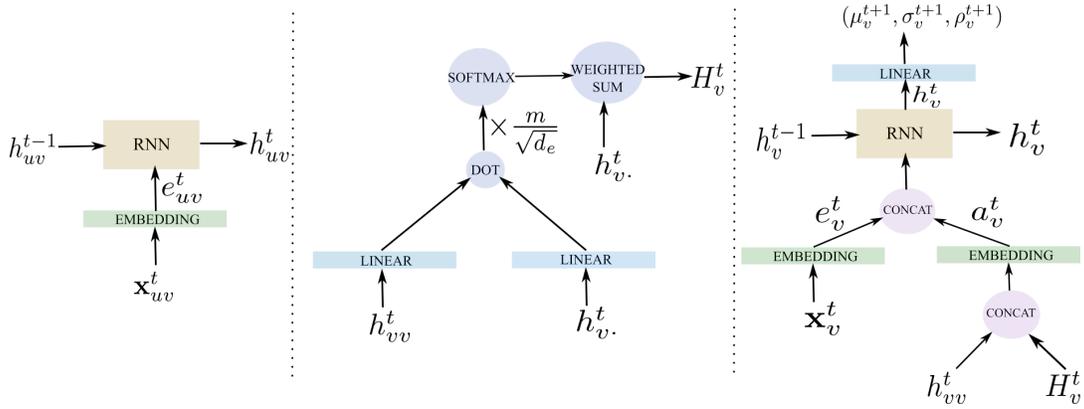}
  \caption{Architecture of EdgeRNN (left), Attention module (middle) and NodeRNN (right)}
  \label{fig:arch}
  \vspace*{-0.5cm}
\end{figure*}

Our architecture differs from the S-RNN architecture, by introducing an attention module to compute a soft attention over hidden states of neighboring spatial edgeRNNs for each node
as summarized in Figure~\ref{fig:arch}. We will describe each of these components in the following subsections.

\subsubsection{EdgeRNN}
\label{sec:edgernn}

Each spatial edgeRNN $\rnn_{uv}$, at every time-step $t$, takes the corresponding edge's features $\feature_{uv}^t$, embeds it into a fixed-length vector $e_{uv}^t$ and is used as an input to the RNN cell as follows:
\begin{align}
  e_{uv}^t &= \phi(\feature_{uv}^t; W^e_{\spatial}) \\
  h_{uv}^t &= \RNN(h_{uv}^{t-1}, e_{uv}^t; W_{\spatial}^r)
\end{align}
where $\phi(\cdot)$ is an embedding function, $W_{\spatial}^e$ is the embedding weights, $h_{uv}^t$ is the hidden state of the RNN at time $t$ and $W_{\spatial}^r$ are the weights of the spatial edgeRNN cell.

The temporal edgeRNN $\rnn_{uu}$ is defined in a similar way with its own set of weights $W_{\temporal}^e$ and $W_{\temporal}^r$ for the embedding and edgeRNN, respectively. Hence, the trainable parameters for edgeRNNs are $W_{\temporal} = \{W_{\temporal}^e, W_{\temporal}^r\}$ and $W_{\spatial} = \{W_{\spatial}^e, W_{\spatial}^r\}$.

\subsubsection{Attention Module}
\label{sec:attention-module}

For each node $v$, the attention module computes a soft attention over the hidden states $h_{v\cdot}^t$ of the edgeRNNs $\rnn_{v\cdot}$ of the spatial edges that the node $v$ belongs to. Observe that this differs from the S-RNN architecture from~\cite{jain2016structural}, where the edge features of these spatial edges are added and sent to the edgeRNN to compute a single hidden state, which is used as an input to the nodeRNN.

At each time-step $t$ for each node $v$, we compute a score between the hidden state $h_{vv}^t$ of its corresponding temporal edgeRNN $\rnn_{vv}$ and all the hidden states $h_{v\cdot}^t$ of the neighboring spatial edgeRNNs $\rnn_{v\cdot}$. The score function used is \textit{scaled dot product attention} \cite{2017arXiv170603762V}, given by:
\begin{align}
  \text{score}(h_{vv}^t, h_{v\cdot}^t) = \frac{m}{\sqrt{d_e}} \left< W_1h_{vv}^t, W_2h_{v\cdot}^t\right>
  \label{eq:attn}
\end{align}
where $m$ is the number of spatial edges the node is associated with, $W_1, W_2$ are weights to linearly scale and project the hidden states into $d_e$ dimensional vectors. Scaling the dot product using $\frac{m}{\sqrt{d_e}}$ is necessary because dot product attention performs poorly for large values of $d_e$ as found in~\cite{2017arXiv170603762V}, and the number of spatial edges change from frame to frame, depending on the number of agents.

The output vector $H_v^t$ is computed as a weighted sum of $h_{v\cdot}^t$ with the weights as softmax of computed scores,
\begin{align}
  H_{v}^t = \sum_{i=1}^m \frac{\exp(\text{score}(h_{vv}^t, h_{vi}^t))}{\sum_{j=1}^m \exp(\text{score}(h_{vv}^t, h_{vj}^t))} \cdot h_{vi}^t.
  \label{eq:weights}
\end{align}

Hence, the trainable parameters in the attention module are the weights $W_1$ and $W_2$.

\subsubsection{NodeRNN}
\label{sec:nodernn}

Finally, the nodeRNN $\rnn_v$ at every time-step $t$, takes the corresponding node's features $\feature_v^t$, embeds it into a fixed-length vector $e_v^t$. It also takes the hidden state $h_{vv}^t$ of corresponding temporal edgeRNN $\rnn_{vv}$, concatenates it with the computed attention output $H_v^t$ and embeds it into a fixed-length vector $a_v^t$. These embeddings are concatenated and sent as an input to the RNN cell as follows:
\begin{align}
  e_v^t &= \phi(\feature_v^t; W_{\node}^e) \\
  a_v^t &= \phi(\concat(h_{vv}^t, H_v^t); W_{\node}^h) \\
  h_v^t &= \RNN(h_v^{t-1}, \concat(e_v^t, a_v^t); W_{\node}^r)
\end{align}

The hidden state of the RNN cell at time-step $t$ is passed through a linear layer $W_{\node}^o$ to get a 5D vector $(\mu_v^{t+1}, \sigma_v^{t+1}, \rho_v^{t+1})$ corresponding to predicted mean, standard deviation and correlation of a bivariate Gaussian distribution, similar to~\cite{graves2013generating}.
\begin{align}
  (\mu_v^{t+1}, \sigma_v^{t+1}, \rho_v^{t+1}) = W_{\node}^o h_v^t
\end{align}

Thus, the trainable parameters for a nodeRNN are $W_{\node} = \{W_{\node}^e, W_{\node}^h, W_{\node}^r, W_{\node}^o\}$.

\subsection{Training the model}
\label{sec:training-model}

We jointly train the entire model by minimizing the negative log-likelihood loss {$L_v$} of the node's true position $(x_v^t, y_v^t)$ at all predicted time-steps $t = T_{\obs} + 1, \cdots, T_{\pred}$ under the predicted bivariate Gaussian distribution $(\mu_v^t, \sigma_v^t, \rho_v^t)$ as follows:
\begin{align*}
  L_v(W_{\node}, &W_{\spatial}, W_{\temporal}, W_1, W_2) = \\
                 &- \sum_{t=T_{\obs}+1}^{T_{\pred}} \log(\prob(x_v^t, y_v^t | \mu_v^t, \sigma_v^t, \rho_v^t))
\end{align*}
The loss is computed over trajectories of all nodes in the training dataset and backpropagated. Note that, we jointly backpropagate through the nodeRNN, spatial edgeRNN and temporal edgeRNN, thereby updating all of their parameters to minimize the loss.

\subsection{Inference for path prediction}
\label{sec:infer-path-pred}

At test time, we fit the trained model to observed trajectory at time-steps $t = 1, \cdots, T_{\obs}$ and sample from the predicted bivariate Gaussian distribution to get forecasted locations $\{(\xhat_v^t, \yhat_v^t)\}$ for all the pedestrians, for time-steps $t = T_{\obs}+1, \cdots, T_{\pred}$. {Formally,}
\begin{align}
  (\xhat_v^t, \yhat_v^t) \sim \mathcal{N}(\mu_v^t, \sigma_v^t, \rho_v^t)
\end{align}
For time-steps $t > T_{\obs}+1$, we use the predicted location at the previous time-step $\{(\xhat_v^{t}, \yhat_v^{t})\}$  in-place of the true coordinates $\{(x_v^{t}, y_v^{t})\}$ as node features $\feature_v^t$, similar to~\cite{Alahi-2016-ID2}. The predicted locations are also used to compute the edge features $\feature_{uv}^t$ for these time-steps.


\section{Evaluation}
\label{sec:evaluation}

\subsection{Datasets and Metrics}
\label{sec:datasets-metrics}

We evaluate our model, which we call \textit{Social Attention}, on two publicly available datasets: ETH~\cite{pellegrini09}, and UCY~\cite{lerner07}. These two datasets contain $5$ crowd sets with a total of $1536$ pedestrians exhibiting complex interactions such as walking together, groups crossing each other, joint collision avoidance and nonlinear trajectories, as shown in~\cite{pellegrini09}. These datasets are recorded at $25$ frames per second, annotated every $0.4$ second and contain $4$ different scenes. {As shown in \cite{Alahi-2016-ID2}, Social LSTM performs better than other traditional methods such as linear model, the Social forces model \cite{helbing95} and Interacting Gaussian Processes \cite{trautman10}. Hence, we chose Social LSTM as the baseline to compare the performance of our method.}

To compute the prediction error, we consider the following two metrics:
\begin{enumerate}
\item \textit{Average Displacement Error:} Similar to the metric used in~\cite{pellegrini09}, this computes the mean euclidean distance over all estimated points at each time-step in the predicted trajectory and true trajectory.
\item \textit{Final Displacement Error:} Introduced in~\cite{Alahi-2016-ID2}, this metric computes the mean euclidean distance between the final predicted location and the final true location after $T_{\pred}$ time-steps.
\end{enumerate}

Similar to~\cite{Alahi-2016-ID2}, we use a leave-one-out approach where we train and validate our approach on $4$ sets, and test on the remaining set. We repeat this for all the $5$ sets. For validation, within each set we divide the set of trajectories in a $80-20$ split for training and validation data. Our baseline, Social-LSTM,~\cite{Alahi-2016-ID2}, has also been trained in the same fashion. We observe the trajectory for $T_{\obs} = 8$ time-steps (corresponding to $3.2$ seconds) and predict the trajectory for the next $T_{\pred} - T_{\obs} = 12$ time-steps (corresponding to $4.8$ seconds). We also conduct the same experiments for an independent LSTM approach that models each trajectory independently. 

\subsection{Implementation Details}
\label{sec:impl-deta}

We use LSTM as the RNN in our Social Attention model. The dimension of
hidden state of nodeRNN is set to $128$ and that of edgeRNN to
$256$. All the embedding layers in the network embed the input into a
$64$ dimensional vector with ReLU nonlinearity. The attention
dimension, i.e., $d_e$ in Equation~\ref{eq:attn}, is set to $64$. A
batch size of $8$ is used and the network is trained for $100$ epochs
using Adam with an initial learning rate of $0.001$. The global norm
of gradients are clipped at a value of $10$ to ensure stable
training. The model was trained on a single Titan-X GPU.


\subsection{Quantitative Results}
\label{sec:quantitative-results}
The prediction errors for all the methods on the 5 crowd sets is
presented in Table \ref{tab:results}. The naive independent LSTM
approach results in high prediction errors, as it cannot capture
human-human interactions unlike Social LSTM and Social
Attention. However, in some cases, the independent LSTM approach
performs slightly better than others, especially in sparse crowd
settings where there are scarcely any interactions. Our model, Social
Attention, performs better than Social LSTM consistently across all
the crowd sets in both the metrics.
\begin{table*}[ht]
  \centering
  \caption{Prediction errors (in metres) on all the crowd sets for all the methods}
  \label{tab:results}
  \begin{tabular}{|l|l|l|l|l|}
    \hline
    \textbf{Metric}                                      & \textbf{Crowd Sets}     & \textbf{LSTM} & \textbf{Social LSTM} & \textbf{Social Attention} \\ \hline
    \multirow{5}{*}{\textbf{Average Displacement Error}} & \texttt{ETH - Univ}   &  0.59   &   0.46          &   0.39               \\ \cline{2-5} 
                                                         & \texttt{ETH - Hotel}  & 0.35    &    0.42         &       0.29           \\ \cline{2-5} 
                                                         & \texttt{UCY - Zara 1} & 0.25    &    0.21         &      0.20            \\ \cline{2-5}
                                                         & \texttt{UCY - Zara 2} &  0.38   &    0.41         &      0.30            \\ \cline{2-5}
                                                         & \texttt{UCY - Univ}   &  0.40   &     0.36        &       0.33           \\ \cline{2-5}
                                                         & \texttt{Average}   &  0.39  &      0.37       &      \textbf{0.30}            \\ \hline\hline
    \multirow{5}{*}{\textbf{Final Displacement Error}}   & \texttt{ETH - Univ}   & 5.28    &  4.55           &   3.74               \\ \cline{2-5} 
                                                         & \texttt{ETH - Hotel}  & 4.42    &    3.57         &     2.64             \\ \cline{2-5} 
                                                         & \texttt{UCY - Zara 1} & 1.55    &    0.65         &     0.52             \\ \cline{2-5} 
                                                         & \texttt{UCY - Zara 2} & 3.57    &     3.39        &      2.13            \\ \cline{2-5}
                                                         & \texttt{UCY - Univ}   & 6.39    &     4.45        &      3.92            \\ \cline{2-5}
                                                         & \texttt{Average}   &  3.84   &     3.32        &    \textbf{2.59}              \\ \hline
  \end{tabular}
\end{table*}
{In particular, in the \texttt{ETH-Hotel} crowd set, our approach significantly outperforms others by a large margin, supporting our hypothesis on non-local interactions as follows}. This crowd set contains a lot of pedestrians who are stationary or go towards each other with varied velocities and heading. For stationary pedestrians, Social LSTM considers them important if they are within the local neighborhood, whereas Social Attention does not assign importance to these agents as they don't affect others motion in a significant way. In the case of pedestrians going headlong towards each other, Social LSTM does not consider them until they enter each others local neighborhood, whereas Social Attention captures the interactions between them from a far distance based on their velocities and heading. By learning relative importance of each pedestrian in the crowd from data, Social Attention results in more accurate predictions.

In our evaluation, we also included the prediction errors of pedestrians for whom we observed fewer than $T_{\obs}$ time-steps as they entered the crowd at a later time. Generally, when we have a fewer number of observations, the model's accuracy naturally degrades for their predictions. This is one of the primary reasons for the difference in our results of Social LSTM compared to that from the original paper~\cite{Alahi-2016-ID2}, as they disregarded such scenarios. On the other hand, we consider them to be important since they happen often in real robot navigation.

{ Accounting for all agents in the crowd increases the computational complexity of our approach, but inference in the model is parallelized on GPU to ensure real-time performance (10Hz)}.

\subsection{Qualitative Results}
\label{sec:qualitative-results}

\begin{figure*}
  \centering
  \includegraphics[width=0.7\linewidth]{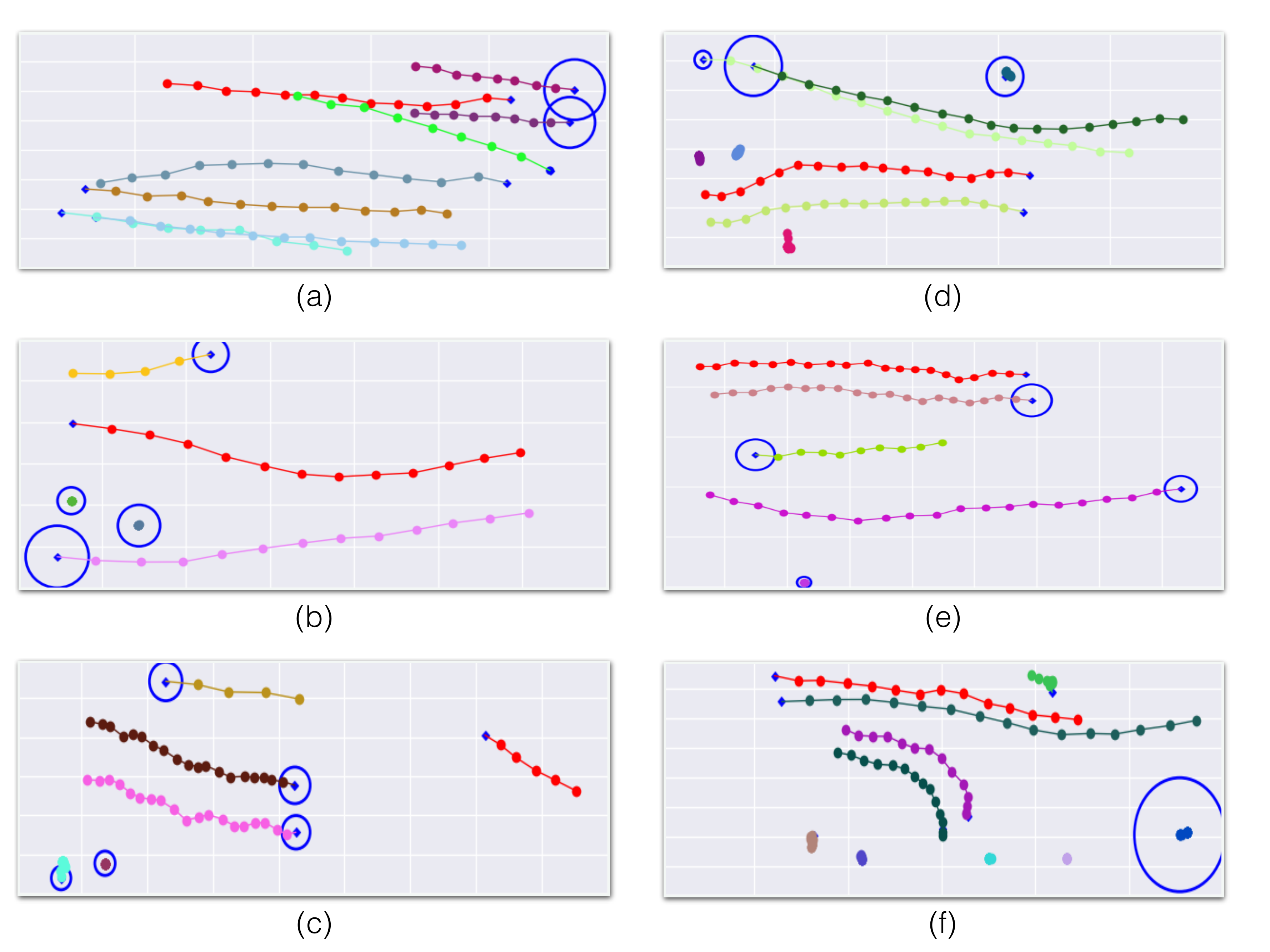}
  \caption{Attention weights predicted by Social Attention. In all plots, the red trajectory is the pedestrian whose attention weights are being predicted. The rest of the trajectories are other pedestrians in the crowd. Solid dots represent locations of the pedestrians at previous time-steps and the blue diamond marker represents their current locations. The circles around the current position of surrounding pedestrians represents attention with the radius proportional to attention weight.}
  \label{fig:results}
  \vspace*{-0.5cm}
\end{figure*}

The qualitative results for Social Attention is shown in Figure \ref{fig:results}. To analyze the learned attention model, we considered several crowd scenarios among the datasets and extracted the predicted attention weights (softmax of scores in equation \ref{eq:weights}). This lets us observe the relative importance of each pedestrian on the motion of a specific pedestrian, as predicted by Social Attention. Figure \ref{fig:results} {(a)-(c)} show scenarios where the model successfully identifies important pedestrians and Figure \ref{fig:results} {(d)-(f)} highlight the scenarios where the model fails. In (a), the model attends with a higher weight to the dynamic pedestrian in close proximity compared to others far away. (b) shows a scenario where the model predicts that stationary pedestrians in the local neighborhood are relatively less important than a dynamic pedestrian who is farther away. In (c), the model assigns equal relative importance to each of the dynamic pedestrians as they are all too far away to exert any influence.

There are several cases where our model incorrectly predicts the relative importance. Figure~\ref{fig:results} (d) and (f) show those scenarios where the model assigns a high attention weight to pedestrians who are far and moving in such a way (or stationary as in (f)) that they can't exert any influence, completely ignoring nearby pedestrians who are more important. Finally in (e), the model predicts equal attention weights for all the three dynamic pedestrians, while one of them is clearly more important than others. { Investigating the reason for such prediction failures of our model is left to future work.}

\section{Conclusion}
\label{sec:conclusion}

Identifying the need for accurate human trajectory prediction 
for autonomous robot navigation in crowds, we have presented an
attention-based trajectory
prediction model, \textit{Social Attention}. Our model learns the relative
influence of each pedestrian in the crowd on the planning behavior of
the other, and accurately predicts their future trajectories. We use
an RNN mixture to model both the temporal and spatial dynamics of
trajectories in human crowds. The resulting model is feedforward,
fully-differentiable, and is jointly trained to capture human-human
interactions between pedestrians. We show that our proposed method
outperforms the state-of-the-art approach in prediction errors, on two
publicly available datasets. We also analyze the learned attention
model to understand which surrounding agents humans attend to, when
navigating a crowd, and present qualitative results. Future work can
extend the model to include static obstacles in the environment. The
S-RNN architecture employed in this work can be naturally extended to
model different semantic entities, as shown
in~\cite{jain2016structural}. We also plan to verify and validate our
model on a real robot placed in a human crowd, predicting future
trajectories of surrounding humans and planning its own path (using an
approach like \cite{vemula2016path}) to reach its destination. {In addition to these, it would be useful to compare performance of our model with IRL-based approaches such as \cite{kuderer12, kretzschmar16, pfeiffer16}, which currently don't scale well to large crowds.}


\small
\section*{ACKNOWLEDGMENTS}
This work was conducted in part through
collaborative participation in the Robotics Consortium sponsored by
the U.S Army Research Laboratory under the Collaborative Technology
Alliance Program, Cooperative Agreement W911NF-10-2-0016. The views
and conclusions contained in this document are those of the authors
and should not be interpreted as representing the official policies,
either expressed or implied, of the Army Research Laboratory of the
U.S. Government. The U.S. Government is authorized to reproduce and
distribute reprints for Government purposes notwithstanding any
copyright notation herein.


\bibliographystyle{IEEEtranS}
\bibliography{ref}  

\end{document}